\documentclass[12pt]{amsart}
\usepackage{geometry}                
\usepackage{graphicx}
\usepackage{amssymb}
\usepackage{epstopdf}
\usepackage{url}
\usepackage[toc,page]{appendix}
\usepackage{vmargin}
\setmarginsrb{25mm}{10mm}
             {25mm}{10mm}
             {10mm}{10mm}
             {10mm}{12mm}

\usepackage{color}

\title[Instructional Texts for Plan Generation]{Towards Evaluating Plan Generation Approaches with Instructional Texts}
\author[Chowdhury et al.]{Debajyoti Paul Chowdhury$^{1}$, Arghya Biswas$^{1}$, Tomasz Sosnowski$^{1}$, Kristina Yordanova$^{1,2,3}$}

\begin{document}
\maketitle

\begin{tabular}{p{0.4\textwidth} p{0.6\textwidth}l}
\begin{flushleft}
\includegraphics[width=0.35\textwidth]{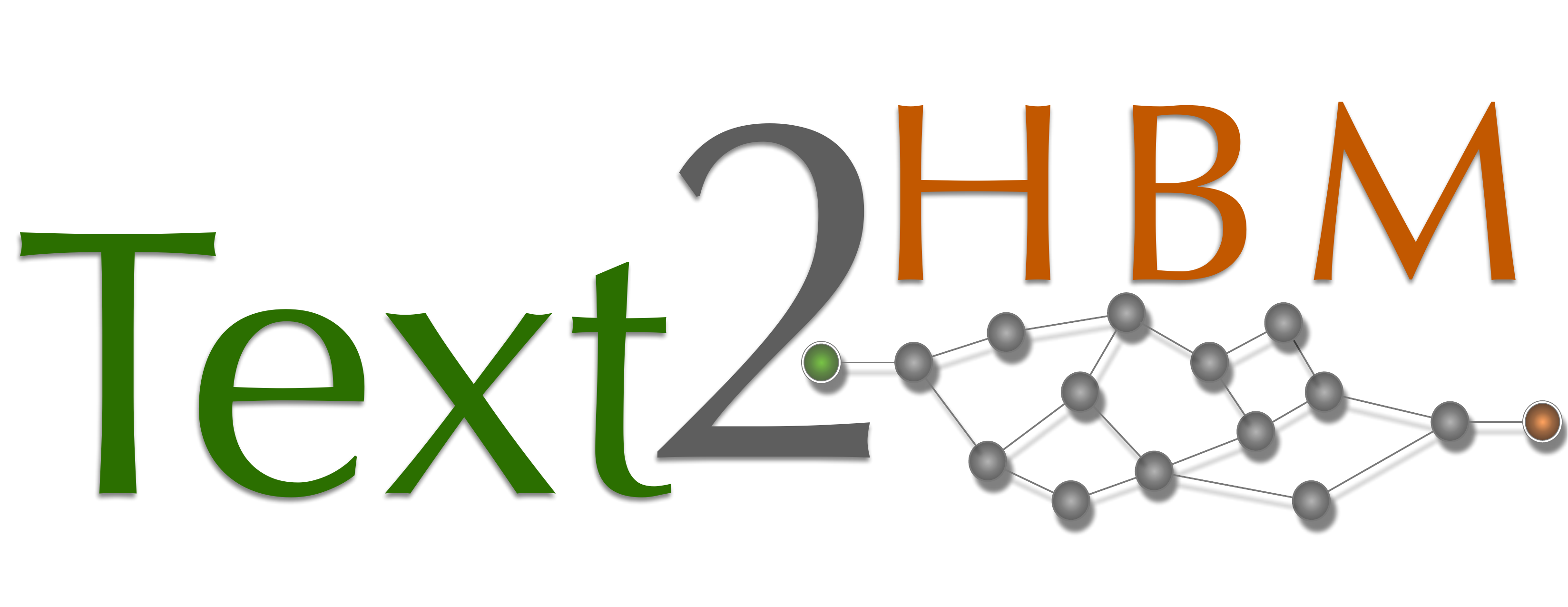}
\end{flushleft}
&
\begin{flushleft}
$^1$ University of Rostock, Rostock, Germany\\
$^2$ University of Bristol, Bristol, UK\\
$^3$ Kurt Lewin Center for Theoretical Psychology\\
\end{flushleft}\\
\end{tabular}

\begin{abstract}
Recent research in behavior understanding through language grounding has shown it is possible to automatically generate behaviour models from textual instructions. 
These models usually have goal-oriented structure and are modelled with different formalisms from the planning domain such as the Planning Domain Definition Language. 
One major problem that still remains is that there are no benchmark datasets for comparing the different model generation approaches, as each approach is usually evaluated on domain-specific application. 
To allow the objective comparison of different methods for model generation from textual instructions, in this report we introduce a dataset consisting of 83 textual instructions in English language, their refinement in a more structured form as well as manually developed plans for each of the instructions. 
The dataset is publicly available to the community. 

\end{abstract}

\section{Introduction}

Intelligent assistive systems support the daily activities and allow healthy people as well as people with impairments to continue their independent life \cite{Hoey:2010}. 
To provide timely and adequate support, such systems have to recognise the user actions and intentions, track the user interactions with objects, detect errors in the user behaviour, and find the best way of assisting them  \cite{Hoey:2010}. 
This can be done by activity recognition (AR) approaches that utilise human behaviour models (HBM) in the form of rules. 
These rules are used to generate probabilistic models with which the system can infer the user actions and goals \cite{Krueger:2014,Trafton:actr-e,Hiatt,geffner,Baker.ea:2009}. 
This type of models is also known as computational state space models (CSSM) \cite{Krueger:2014}. 
CSSMs perform activity recognition by treating it as a plan recognition problem. 
In plan recognition given an initial state, a set of possible actions, and a set of observations, the executed actions and the user goals have to be recognised \cite{Ramirez:2009}.  
CSSMs use prior knowledge to obtain the context information needed for building the user actions and the problem domain. 
The prior knowledge is provided in the form of precondition-effect rules by a domain expert or by the model designer. 
This knowledge is then used to manually build a CSSM.
The manual modelling is, however, time consuming and error prone process \cite{Nguyen:2013,problemsKrueger}.

To address this problem, different works propose to learn the models from sensor data \cite{Zhuo.Kambhampati:2013,Okeyo:2011}.
One problem these approaches face is that sensor data is expensive \cite{Ye:2014}. 
Another problem is that sensors are not always able to capture fine-grained activities \cite{sensorsGranularity:Chen}, thus, they might potentially not be learned. 

To reduce the impact of domain experts or sensor data on the model performance, one can substitute them with textual data \cite{Philipose:2004}.
In other words, one can utilise the knowledge encoded in textual instructions to learn the model structure.
Textual instructions specify tasks for achieving a given goal or performing a certain task without explicitly stating all the required steps. 
This omission of certain steps makes them a challenging source for learning a model \cite{Branavan:2010}.  
On the positive side, they are usually written in imperative form, have a simple sentence structure, and are highly organised.
Compared to rich texts, this makes them a better source for identifying the sequence of actions needed for reaching the goal \cite{Zhang:2012}.

There are different works that address the problem of learning models from textual instructions.
Such models are used for constructing plans of human behaviour \cite{Li:2010,Branavan:2012,Yordanova:2018KI,Yordanova:2016a}, for learning an optimal actions' execution sequence based on natural instructions \cite{Branavan:2011,Branavan:2010,Vogel:2010,Chen:2011,Babes-Vroman:2012,Benotti:2014}, for constructing machine understandable model from natural language instructions \cite{Zhang:2012,Kollar:2014}, and for automatically generating semantic annotation for sensor datasets \cite{Yordanova:2020a,Yordanova:2018a}.  
Model learning from textual instructions has applications in different fields of computer science: constructing plans of terrorist attacks \cite{Li:2010}, improving tasks execution (such as navigation, computer commands following, games playing) by interpreting natural language instructions \cite{Branavan:2011,Branavan:2010,Vogel:2010,Chen:2011,Babes-Vroman:2012,Benotti:2014}, the ability of a robot or machine to interpret instructions given in natural language \cite{Zhang:2012,Kollar:2014}, or for behaviour analysis tasks based on sensor observations \cite{Yordanova:2018KI,Yordanova:2017KE}.  

According to \cite{Branavan:2012}, to learn a model of human behaviour from textual instructions, the system has to: 
\begin{enumerate} 
\item \textbf{extract the actions' semantics} from the text, 
\item \textbf{learn the model semantics} through language grounding, 
\item and, finally, to \textbf{translate it into computational model of human behaviour} for planning problems.
\end{enumerate} 
In addition to these requirements, \cite{Yordanova:2016b} introduces 4. the need of \textbf{learning the domain ontology} that is used to abstract and / or specialise the model.

One general challenge that remains is how to empirically compare the different approaches. 
Yordanova \cite{Yordanova:2018KI} proposes different measures such as correctness of the identified elements, complexity of the model, model coverage, and similarity to handcrafted models. 
To compare different approaches with these metrics, however, one needs a common dataset\footnote{Providing datasets and the corresponding annotation is a problem in the community that is often underestimated but that could have negative effect on the overall performance of the system \cite{Yordanova:2018Sensors,Yordanova:2018Arduous,Yordanova:2019c}. In the case of automatically generated models for behaviour analysis the ground truth is a collection of plans that correspond to the provided textual instructions.}. 
In this technical report we address the problem by introducing a dataset, consisting of 83 textual instructions on different topics from our everyday life, together with refinements of the instructions and the corresponding plans. 
The dataset can be used to compare the ability of approaches for model generation from text to learn a model structure able to explain the corresponding plans.   

\section{Procedure for collecting the textual instructions}

The dataset was collected within the context of the project Text2HBM\footnote{https://text2hbm.org/}. 
The goal of the project is to develop methods for the automatic generation of planning operators and behaviour models from textual instructions.
 
To collect the dataset, we first decided on different topics from our everyday life, for which the instructions are to be collected. 
The topics include cooking recipes, buying movie tickets, changing elements of house appliances, deciding on vacation plan.   
For each type of instructions, we attempted to find different opinions on the execution sequence. 
\begin{figure}[!htb]
      \centering
    \includegraphics[width=0.5\textwidth]{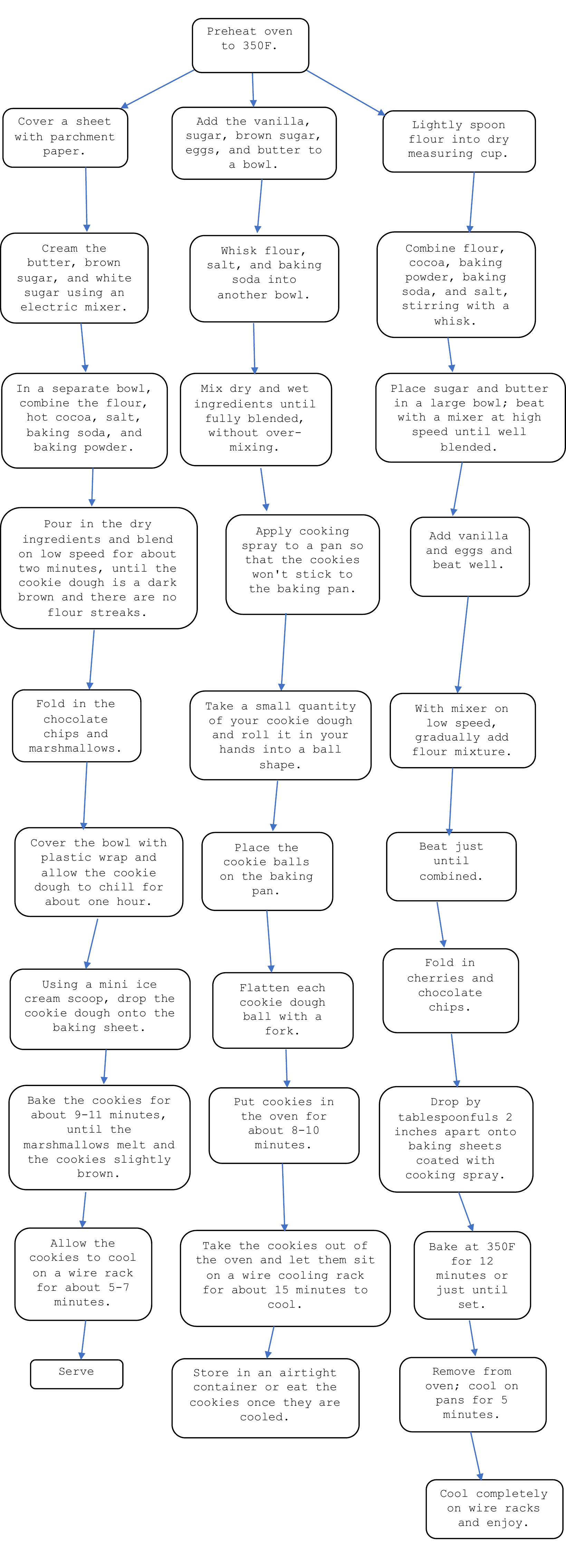}
      \caption{Three different execution sequences or ``recipes'' for preparing chocolate cookies.}
      \label{fig:pop}
     \end{figure}
In other words, we collected the descriptions from different sources addressing the same process. 

 Concretely, we first selected a topic (e.g. – how to make chocolate cookies, how to make a cake etc.), then we asked different people to provide their way of achieving the objective. 
For example, suppose we would like to make ‘chocolate cookies’. 
     We took three different opinions about the making process. 
The three processes are depicted in Figure \ref{fig:pop}.
For each type of instruction, we have collected between one and 14 instructions to cover different variations of the same activity. 
The originally collected texts are our starting point or ``raw data''. 
For example, Figure \ref{fig:instructionRaw} shows the original description collected for preparing chocolate cookies. 
\begin{figure}[!htb]
\centering\scriptsize
\begin{verbatim}
Preheat oven to 350F.
Cover a baking sheet with a parchment paper.
Cream the butter, brown sugar, white sugar using an electric mixer.
In a separate bowl, combine the dry ingredients. Combine the flour, hot cocoa, 
salt, baking soda, and baking powder.
Pour in the dry ingredients and blend on low speed for about two minutes, until 
the cookie dough is a dark brown and there are no flour streaks.
Fold in the chocolate chips and marshmallows.
Cover the bowl with plastic wrap and allow the cookie dough to chill for about one hour.
Using a mini ice cream scoop, drop the cookie dough onto the baking sheet.
Bake the cookies for about 9-11 minutes, until the marshmallows melt and the 
cookies slightly brown.
Allow the cookies to cool on a wire rack for about 5-7 minutes.
Serve.
\end{verbatim}
  \caption{The originally collected text for preparing chocolate cookies.}
      \label{fig:instructionRaw}
\end{figure}   

After initial tests with approaches for model generation we found out that this form of action sentences is not suitable for generating models as each sentence is too complex. 
To address this problem, we performed a refinement step where for each instruction, the sentence was simplified. 
For any sentence, which consists of more than one verb or one verb with many objects to which the action is applied, we broke it into a single verb-object per sentence. 
In other words, we refined the texts so that there is only one action in each step.

\begin{figure}[!htb]
\centering\scriptsize
\begin{verbatim}
Heat the oven to 350F.
Cover the baking sheet with a paper.
Take the electric mixer.
Cream the butter.
Cream the brown sugar.
Cream the white sugar. 
Take a separate bowl.
Combine the dry ingredients.
Combine the flour.
Combine the hot cocoa.
Combine the salt.
Combine the baking soda.
Combine the baking powder.
Pour the dry ingredients.
Blend for two minutes.
Fold the chocolate chips.
Fold the marshmallows.
Cover the bowl with plastic wrap.
Chill the cookie dough.
Put the cookie dough on sheet.
Bake the cookies for 10 minutes.
Cool the cookies for 6 minutes.
Serve.
\end{verbatim}
  \caption{The text for preparing chocolate cookies after refinement.}
      \label{fig:instructionRefined}
\end{figure}   
In Figure \ref{fig:instructionRaw} one can see there are multiple objects, on which the same action is applied (for example, the action ``cream'' is applied to the objects ``brown sugar'', ``white sugar'', and ``butter''). 
To refine it, for each sentence, we broke down the entire sentence into one action (verb) per sentence (see Figure \ref{fig:instructionRefined}). 
We could have created a more structured representation of the steps by removing the articles ``a'' and ``the'', but by doing so an approach relying on natural sentences will probably not be able to parse the sentences correctly.

After refining the texts, we now needed a mechanism for validating the correctness of the generated models.
To achieve that, we manually created plans that represent every ``refined text'' instruction. 
The format for the plans can be seen in Figure \ref{fig:plan}.

\begin{figure}[!htb]
\centering\scriptsize
\begin{verbatim}
0,*,INITIALIZE
1,*,(heat oven F)
2,*,(cover sheet)
3,*,(take mixer)
4,*,(cream butter)
5,*,(cream sugar)
6,*,(cream sugar)
7,*,(take bowl)
8,*,(combine ingredients)
9,*,(combine flour)
10,*,(combine cocoa)
11,*,(combine salt)
12,*,(combine soda)
13,*,(combine powder)
14,*,(pour ingredients)
15,*,(blend)
16,*,(fold chips)
17,*,(fold marshmallows)
18,*,(cover bowl plastic)
19,*,(chill dough)
20,*,(put dough sheet)
21,*,(bake cookies)
22,*,(cool cookies rack)
23,*,(serve)
24,*,FINISHED
\end{verbatim}
  \caption{The plan corresponding to the refined instruction for preparing cookies.}
      \label{fig:plan}
\end{figure}     
Here the first element is the time at which the action starts\footnote{In this case we assume instantaneous actions, such that each action takes one time step. It is however, also possible to have actions with real durations. For example see \cite{Yordanova:2019a}.}, the second element is the $*$, which tells us whether this is a new action or whether it was transferred from the previous step\footnote{This parameter is redundant in single agent behaviour, such as in our case. There are, however, applications where the behaviour of multiple agents is tracked. In such cases, when a new action starts for one agent, it is possible that the action for another agent is just transferred from the previous one. For example, see \cite{icaart,Yordanova:2014}.}, and the third element shows the concrete step of the plan in the form of action followed by objects (entities), on which the action is executed.

It can be seen, that we followed a certain structure when defining the plan steps. 
After initialising, the next steps are written in a format similar to the structure of the refined sentences. 
Each step starts with a verb, followed by the primary object and succeeded by the secondary object. 
Usually, the secondary object is the source or destination location of the primary object.
For example, \textbf{cream (verb)   butter (primary object)   mixer (secondary object)}.
More generally, the form of the steps follows the formula
\textbf{N,*,(action object1/location1 object2/location2)}.

\section{Dataset}

In the following we describe the data format. 
The dataset is publicly available and can be downloaded from \url{https://github.com/stenialo/Text2HBM}. 

\subsection{Data format}

The entire dataset consists of various activities resulting in 83 unique descriptions. 
It is separated in eleven categories: Cake, Chicken, Filter system, Movie Tickets, PC cooler, Portable AC, Vacation Plan, Cookies, French Toasts, Sushi, Washing Machine.

In each of these activities there are certain types of procedures which are categorised in terms of different names. 
\begin{itemize}
\item \textbf{Cookies} has fourteen different procedures for making some kinds of cookies.
\item \textbf{French Toast} has fourteen different procedures for making some kinds of French toasts.
\item \textbf{Sushi} has thirteen different procedures for making some kinds of sushi.
\item \textbf{Washing Machine} has five different procedures for cleaning and other stuffs related to washing machine.
\item \textbf{Vacation Plans} has eight different procedures for making some kind of trip or vacation.
\item \textbf{Cake} has ten different opinions from ten different persons for procedures of making cake.
\item \textbf{Chicken} also has ten different opinions from ten different persons for procedures of making cake.
\item \textbf{Portable Air Conditioner} has four different processes for installing.
\item \textbf{Movie Ticket} booking has two different options.
\item \textbf{House Water-Filter System} also has two different opinions.
\item \textbf{PC Cooling System} has one procedure.
\end{itemize}

Each of these procedures/opinions/processes has three types of data. 
These three categories are: 
\begin{itemize}
\item \textbf{Raw Data:} As the term suggests, it is the raw texts collected from different people for different kind of preparation in each of the activities mentioned. The sentences can be complex or simple depending on the amount of parts of speech in the particular context.
\item \textbf{Refined Text:} To make the Raw Data easily processable, the sentences of the steps in Raw Data needed to be simplified. This form basically starts with a main verb followed by the main object and ended by the secondary object. Usually, the last term happens to be the location of the object. But sometimes there can be cases where the location is not important. But each and every step has to follow the structure of ‘single verb and single primary object’. Also, each sentence has to be grammatically correct and meaningful. For this reason, all the required prepositions and articles can not be avoided.
\item \textbf{Plans:} To evaluate the  generated models we needed particular plan data for each of the corresponding texts. After initialising the format looks like
\textbf{N,*,(action object1/location1 object2/location2)}
\end{itemize}

\subsection{Folder structure}

The dataset has the following folder structure.
\begin{itemize}
\item Textual\_Descriptions
	\begin{itemize}
	\item Plans
	\item Raw\_Data
	\item Refined\_Text
	\item Refined\_Text2
	\end{itemize}
\end{itemize}

Folder \textbf{Plans} contains the plans corresponding on the refined text. As the descriptions were collected by two people, in some of the folders there is additional folder \textbf{Refined\_Text2}. If this folder exists for the given instruction, then the plans are based on \textbf{Refined\_Text2}. Each step in a plan has the form \textbf{Time,*,(action object object)} (e.g. 1,*,(take cup table)). The time indicates the start time of the action, * indicates that the action is a new actions, and the concrete action with the involved elements of the environment is in brackets.

Folder \textbf{Raw\_Data} contains the original instructions.

Folder \textbf{Refined\_Text} contains a refinement of the original instructions from folder \textbf{Raw\_Data}. This refinement consists of making the sentences shorter, with only one verb per sentence. The sentences also start with a verb in imperative form.

Folder \textbf{Refined\_Text2} contains a refinement of the texts in folder \textbf{Refined\_Text}. As the data was collected by two persons, for some instructions the intermediate refinement was not kept, thus the final refinement is in folder \textbf{Refined\_Text}.

The dataset can be downloaded from the github repository \url{https://github.com/stenialo/Text2HBM}.

\section{Conclusion}

In this work we reported a dataset for evaluating automatically generated planning models. 
These models are generated from textual instructions and can be used for various applications.
The aim of the dataset is to provide a benchmark for approaches that attempt to learn planning operators from instructional texts. 
The manually built plans allow to test the validity of each generated model against the corresponding plan or even against other plans describing the same activity. 
We hope that this dataset will be the first step towards identifying common metrics and data for evaluating methods for automatic generation of behaviour models.   

\section*{Acknowledgements}
This work is part of the Text2HBM project, funded by the German Research Foundation (DFG), grant number YO 226/1-1.

\bibliographystyle{abbrv}
\bibliography{ref}

\end{document}